\documentclass[letterpaper, 10 pt, conference]{ieeeconf}
\IEEEoverridecommandlockouts                            
\usepackage{amsmath,amsfonts}
\usepackage{algorithmic}
\usepackage{algorithm}
\usepackage{array}
\usepackage[caption=false,font=footnotesize,labelfont=rm,textfont=rm]{subfig}
\usepackage{textcomp}
\usepackage{stfloats}
\usepackage{url}
\usepackage{verbatim}
\usepackage{graphicx}
\usepackage{cite}
\usepackage{color}
\usepackage{booktabs}
\usepackage{multirow}
\usepackage{geometry}
\begin{document}

\title{\LARGE \bf DACOOP-A: Decentralized Adaptive Cooperative Pursuit via Attention}

\author{Zheng Zhang$^{1}$, Dengyu Zhang$^{1}$, Qingrui Zhang$^{1}$, Wei Pan$^{2}$, Tianjiang Hu$^{1}$
\thanks{ $^{1}$Machine Intelligence and Collective Robotics (MICRO) Lab, Sun Yat-sen University, Shenzhen, China (\tt\{{\tt\small
 zhangzh363, zhangdy56}\}@mail2.sysu.edu.cn; \{\tt{\small zhangqr9, hutj3}\}@ 
 mail.sysu.edu.cn).}
\thanks{$^{2}$Department of Computer Science, The University of Manchester, UK and Department of Cognitive Robotics, Delft University of Technology, Netherlands \tt\small wei.pan@tudelft.nl}
}

\maketitle

\begin{abstract}
Integrating rule-based policies into reinforcement learning promises to improve data efficiency and generalization in cooperative pursuit problems. However, most implementations do not properly distinguish the influence of neighboring robots in observation embedding or inter-robot interaction rules, leading to information loss and inefficient cooperation. This paper proposes a cooperative pursuit algorithm named Decentralized Adaptive COOperative Pursuit via Attention (DACOOP-A) by empowering reinforcement learning with artificial potential field and attention mechanisms. An attention-based framework is developed to emphasize important neighbors by concurrently integrating the learned attention scores into observation embedding and inter-robot interaction rules. A KL divergence regularization is introduced to alleviate the resultant learning stability issue. Improvements in data efficiency and generalization are demonstrated through numerical simulations. Extensive quantitative analysis and ablation studies are performed to illustrate the advantages of the proposed modules. Real-world experiments are performed to justify the feasibility of deploying DACOOP-A in physical systems. 
\end{abstract}


\section{Introduction}
\begin{figure}
    \centering
    \includegraphics[width=0.48\textwidth]{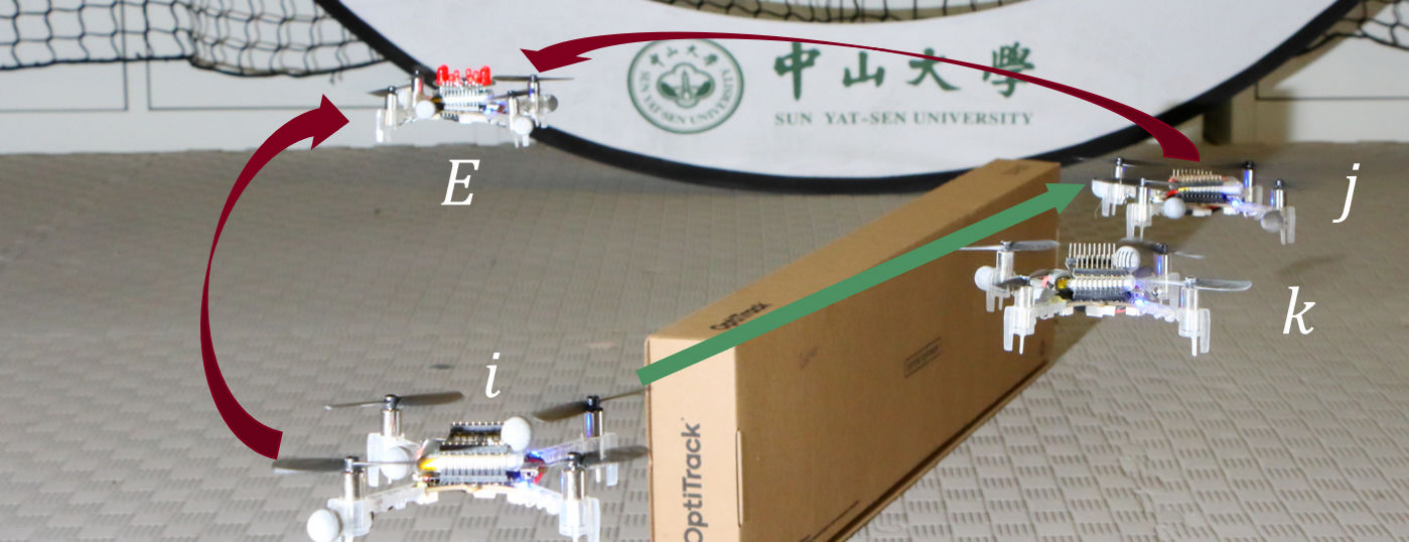}
    \caption{The cooperative pursuit. $E$ is the evader while $i,j,k$ are pursuers. {The red arrows denote the encirclement formed by $i$ and $j$. It implies $j$ chases $E$ from the right side, while $i$ cuts off the escape route of $E$ from the left side.} The green arrow denotes the neighboring robot that $i$ should attend to. $k$ is not attended to because it has less potential to cooperate.}
    \label{fig:fig1}
\end{figure}

Cooperative pursuit aims to coordinate multiple pursuers for capturing one evader in a decentralized manner \cite{chung2011search}, {as shown in Fig. \ref{fig:fig1}}. Most existing algorithms employ manually designed rules with domain knowledge \cite{zhou2016cooperative,fang2020cooperative}. For example, the pursuit-evasion problem with collision avoidance is addressed by combining several forces in \cite{janosov2017group}. However, manually designing cooperative pursuit rules in complicated scenarios is intractable, as robots might encounter numerous environment states. Furthermore, the performance of rule-based methods is sensitive to problem settings and parameter configurations, making them inapplicable in real-world tasks \cite{muro2011wolf}. 

Compared with rule-based methods, reinforcement learning (RL) is more promising for learning sophisticated cooperation because it is possible to obtain various abilities to maximize rewards \cite{silver2021reward,gupta2017cooperative}. However, most RL methods are notorious for the data inefficiency issue that is more severe in multi-robot environments \cite{zhang2021multi}. One of the reasons is an inherent non-stationarity of the environment challenges that value-based RL algorithms. At the same time, policy-based RL methods suffer from a variance that increases as the number of robots \cite{lowe2017multi}. Another challenge for RL is the limited generalization ability. Most RL algorithms focus on maximizing accumulated rewards in predefined training environments. However, the implementation environments commonly have different setups. Such differences would degenerate the performance of the learned policies in real applications \cite{tan2018sim}. 

To improve data efficiency and generalizability, our previous work, DACOOP, introduces a hybrid design that integrates rule-based policies, artificial potential field (APF), into RL for cooperative pursuit \cite{dacoop}. Though DACOOP performs better than vanilla RL algorithms, its performance is still limited. The first reason is that the {Q network} of DACOOP takes the mean embedding of all neighboring robots as input. Since the importance of neighboring robots varies in each state, arbitrary average operation leads to inevitable information loss and data inefficiency. The second reason is that the mean embedding results usually deviate from those in training scenarios once the system size changes, thus deteriorating the generalization capability of the learned policies. The third reason is that APF is inflexible and suboptimal for multi-robot pursuit problems because it considers all neighboring robots equally.

To tackle the aforementioned problems, a Decentralized Adaptive COOperative Pursuit via Attention (DACOOP-A) algorithm is proposed in this paper by enhancing DACOOP with attention. Our first contribution is an attention-based framework that concurrently integrates the learned attention scores into observation embedding and APF. The attention module is first synthesized with the observation encoding to distinguish important neighboring robots. Compared with mean embedding, it can mitigate information loss and exclude unnecessary information, thus leading to improved generalization. Secondly, the learned attention scores are also employed in APF to weigh the influence of neighboring robots in evaluating inter-robot forces, resulting in an artificial potential field with attention (APF-A) method. 

The second contribution of this paper is to improve the learning stability by augmenting standard RL loss functions with a KL divergence regularization. Introducing attention scores in APF-A would make the state transition probability non-stationary in the training process. Hence, the KL divergence regularization is used to penalize foolhardy updates in the outputs of the attention module, which is key to alleviating the non-stationarity issue.

The third contribution is that ablation studies are conducted to show the efficiency of different modules in the proposed algorithm. Extensive quantitative analysis is thereafter performed to illustrate the potential reasons for the advantages of those respective modules. Additionally, the learned policies are deployed directly in physical quadrotor systems to verify the effectiveness of DACOOP-A.

The remainder of this paper is organized as follows. Related works and preliminaries are provided in Sections \ref{sec:related_works} and \ref{sec:preliminaries}. Section \ref{sec:methodology} presents the implementation details of the proposed algorithm. Experiment results are given in Section \ref{sec:results}. Finally, conclusions and future works are available in Section \ref{sec:conclusions}.

\section{Related works} \label{sec:related_works}
Most RL methods solve multi-robot pursuit problems in an independent manner \cite{matignon2012independent,jiang2018learning,de2021decentralized} or following the centralized training decentralized execution paradigm \cite{lowe2017multi,li2019robust}. Although their performance has been proven in various problem settings, the problem of varying numbers of neighboring robots is intractable in partially observable environments because fully connected networks necessitate fixed-length inputs. To address this problem, Hüttenrauch \emph{et al.} proposed mean embedding that averages the embedding of all neighboring robots firstly \cite{huttenrauch2019deep}. To the same end, Everett \emph{et al.}  processed the information of neighboring robots with LSTM \cite{hochreiter1997long} and then {fed} the last hidden state into policy networks \cite{everett2018motion}. However, both mean embedding and LSTM compress the information of neighboring robots regardless of their significance, usually resulting in the loss of significant information.

Attention mechanisms aim to identify significant elements in sequences and have witnessed exciting success in various domains \cite{bahdanau2015neural}. {In the domain of multi-agent RL, Wen \emph{et al.} employed Transformer and multi-agent advantage decomposition theorem to transform the joint policy search problem into a sequential decision-making process \cite{wen2022multi}. To solve the distraction issue and prompt out-of-sight range cooperation, Shao \emph{et al.} introduced an action prediction model to improve the highly contributed attention while compressing the remaining information to brief messages \cite{shao2023complementary}.} Based on the structure of MADDPG, Iqbal \emph{et al.} used attention mechanisms to synthesize observations of neighboring robots before feeding them into centralized critics \cite{iqbal2019actor}. Besides, multi-robot graph attention networks allow robots to focus on communication channels connected to significant neighboring robots {\cite{qi2022cascaded,niu2021multi,niu2022adaptable}}. However, all the aforementioned works only employ attention mechanisms as information processing approaches. In comparison, we additionally use the attention scores to improve robots' predefined behavior rules, promising to facilitate the learning process further.

\section{Problem Formulation} \label{sec:preliminaries}
\subsection{Multi-robot Pursuit Problem}
This paper considers the multi-robot pursuit problem for $N$ robots capturing one faster evader in a confined environment with obstacles. The set of pursuers is denoted by $\mathcal{V}=\{1,2,\cdots,N\}$, and the evader is indicated by $E$. The evader is considered captured by $i \in \mathcal{V}$ if $d_{E,i}<d_c$, where $d_{E,i}$ is the distance between $E$ and $i$. $d_c$ is the capture range. All robots have a safe radius of $d_s$. Collision occurs if $d_{j,i}<2d_s$ or $d_{o,i}<d_s$, where $d_{j,i}$ is the distance between $i$ and $j$. $d_{o,i}$ is the distance between $i$ and the nearest obstacle. Overall, the objective of the pursuers is formulated as follows.
\begin{equation}
\begin{cases}
    d_{E,i}(t_{max})<d_c \\
    d_{o,i}(t)>d_s \\
    d_{j,i}(t)>2d_s
\end{cases}
\forall i,j \in \mathcal{V}, i \neq j, \forall t \in [0,t_{max}],
\end{equation}
where $t_{max}$ is the task horizon. The environment is partially observable. It implies that only neighboring robots within the perception range $d_p$ can be detected by $i$. Denote the azimuth angle of the evader, the nearest obstacle, the neighboring robot $j$ in the local frame of $i$ as $\phi_{E,i}$, $\phi_{o,i}$, and $\phi_{j,i}$, respectively. {The observations of $i$ include the evader information, the nearest obstacle information, and the neighboring robot information, $\{d_{E,i},\phi_{E,i},d_{o,i},\phi_{o,i},\{d_{j,i},\phi_{j,i}\}_j\}$, where $j \in \mathcal{N}_i$ and $\mathcal{N}_i$ is the set of observable pursuers of $i$.}
The first-order point-mass model is assumed for all robots in this paper.
\begin{equation}
    \dot{\boldsymbol{p}}=\boldsymbol{v},
\end{equation}
where $\boldsymbol{p}$ is the position and $\boldsymbol{v}$ is the velocity. All robots are assumed to move at a constant speed, which is $v_P$ for pursuers and $v_E$ for the evader. They only change their heading angles. To make learned policies applicable in physical systems, the steering commands of pursuers are limited to $[-30^\circ,30^\circ]$.

\subsection{Decentralized Adaptive COOperative Pursuit (DACOOP)}
{The multi-robot pursuit problem can be formulated as a partially observable Markov Game (POMG) that is described by $(\mathcal{S},\mathcal{V},\mathcal{A},\mathcal{P},\mathcal{R},\gamma,\mathcal{Z},\mathcal{O})$, where $\mathcal{S}$, $\mathcal{V}$, $\mathcal{A}$, $\mathcal{P}$, $\mathcal{R}$, $\gamma$, $\mathcal{Z}$, $\mathcal{O}$ are global state space, the set of agents, joint action space, state transition probability, reward function, discount factor, observation function, and local observation space, respectively. At each timestep, agent $i$ chooses an action $a_i$ according to its local observations $o_i$ that are sampled from the global state $s$, and then receives its reward $r_i$. The environment state transits according to the joint action from all agents. The objective of each agent is to learn the optimal policy $\pi^*(a_i|o_i)$ that maximizes its own accumulated rewards.}

Our previous work DACOOP employs APF to improve the data efficiency and generalization ability of vanilla RL in multi-robot pursuit problems \cite{dacoop}. Specifically, the APF navigates pursuers through the combination of three predefined forces. For each pursuer $i \in \mathcal{V}$, the attractive force is
\begin{equation}
    \label{eq:attract}
    \boldsymbol{F}_{a,i}=\frac{\boldsymbol{p}_{E}-{\boldsymbol{p}_i}}{d_{E,i}}.
\end{equation} 
The repulsive force $\boldsymbol{F}_{r,i}$ is
\begin{equation}
    \label{eq:repulse}
    \boldsymbol{F}_{r,i}=\begin{cases}
    \eta (\frac{1}{d_{o,i}}-\frac{1}{\rho})\frac{\boldsymbol{p}_i-\boldsymbol{p}_{o,i}}{d^3_{o,i}}, & \text{if } d_{o,i} \leq \rho \\
    \boldsymbol{0}, & \text{if } d_{o,i}>\rho
    \end{cases}
\end{equation}
where $\eta$ is the scale factor and $\rho$ is the influence range of obstacles.
The inter-robot force is 
\begin{equation}
    \label{eq:individual}
    \boldsymbol{F}_{in,i}=\sum_{\substack{j\in \mathcal{N}_i}}(0.5-\frac{\lambda}{d_{j,i}})\frac{\boldsymbol{p}_j-\boldsymbol{p}_i}{d_{j,i}},
\end{equation}
where $\mathcal{N}_i$ is the set of observable pursuers of $i$. $\lambda$ regulates the compactness of the multi-robot system, which is significant for the emergence of cooperation as demonstrated in \cite{dacoop}. The resulting force $\boldsymbol{F}_{i}=\boldsymbol{F}_{a,i}+\boldsymbol{F}_{r,i}+\boldsymbol{F}_{in,i}$ is used to specify the expected heading of $i$. To alleviate the local minima issue, the wall following rules are introduced.

DACOOP uses the classical RL algorithm D3QN \cite{wang2016dueling} to learn a shared optimal policy $\pi^*(\lambda,\eta|s)$ that outputs the optimal parameter pair $(\lambda,\eta)$ for each pursuer in each time step. Each pursuer calculates its resulting force and moves accordingly after receiving $(\lambda,\eta)$.

\section{Methodology} \label{sec:methodology}
The proposed algorithm DACOOP-A follows the fundamental structure of DACOOP. It implies that DACOOP-A employs vanilla RL to learn the optimal parameter pair $(\lambda,\eta)$ at each timestep as usual. However, compared with DACOOP, DACOOP-A introduces an attention module, an artificial potential field with attention (APF-A) method, and a KL divergence regularization to improve the data efficiency and generalization. This section will describe the proposed algorithm from the three aforementioned aspects. 

\begin{figure*}
    \centering
    \includegraphics[width=0.95\textwidth]{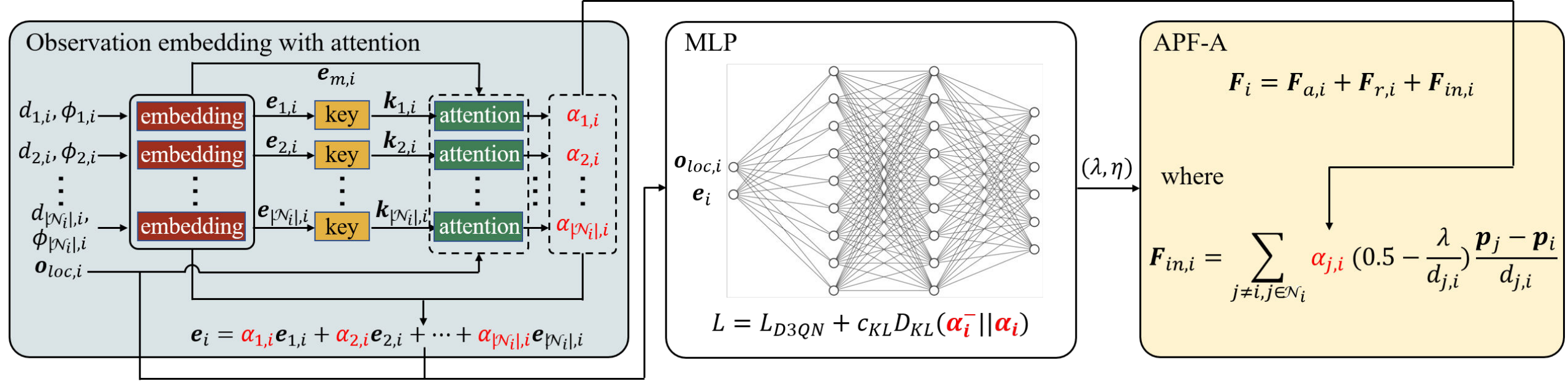}
    \caption{The overview of DACOOP-A. The red, yellow, and green blocks are one-layer fully-connected networks $f_e$, $f_k$, and $f_a$ in Section \ref{sec:embed+attention}, respectively. The information from the neighboring robots is summarized by taking the weighted mean, where the weights are the attention scores $\alpha_{j,i}$. The results $\boldsymbol{e}_i$ are taken as input by an MLP together with the information of the evader and the nearest obstacle $\boldsymbol{o}_{loc,i}$. The {RL policy} outputs parameter pairs $(\lambda,\eta)$ for APF-A that weight the influence of neighboring robots according to the learned attention scores in the computation of inter-robot forces $\boldsymbol{F}_{in,i}$. To alleviate the non-stationarity issue, a KL divergence regularization $D_{KL}(\boldsymbol{\alpha}_i^-||\boldsymbol{\alpha}_i)$ is attached to standard D3QN loss function $L_{D3QN}$.}
    \label{fig:DACOOP_A}
\end{figure*}

\subsection{Observation Embedding with Attention}\label{sec:embed+attention}
Traditional multi-robot pursuit algorithms average the embedding of all neighboring robots to form fixed-length state representations in partially observable environments \cite{huttenrauch2019deep,dacoop}. However, such mean embedding makes the information of significant neighboring robots unrecoverable. Additionally, the mean embedding results are unreliable once the system size changes because the weights of significant information deviate from those in training scenarios. Therefore, an attention module is introduced to process the observation information in the proposed algorithm.

For each pursuer $i \in \mathcal{V}$, DACOOP-A embeds the information of each observable neighboring robot first,
\begin{equation}
    \boldsymbol{e}_{j,i}= f_e(d_{j,i},\phi_{j,i}),
\end{equation}
where $f_e$ is a one-layer fully-connected network. The embedding $\boldsymbol{e}_{j,i}$ is then transformed into the \emph{key} $\boldsymbol{k}_{j,i}$ via another one-layer fully-connected network $f_k$,
\begin{equation}
    \boldsymbol{k}_{j,i}= f_k(\boldsymbol{e}_{j,i}).
\end{equation}
Let $\boldsymbol{o}_{loc,i}=\{d_{E,i},\phi_{E,i},$ $d_{o,i},\phi_{o,i}\}$. The \emph{query}  {consists} of $\boldsymbol{o}_{loc,i}$ and the mean embedding $\boldsymbol{e}_{m,i}$, where
\begin{equation}
    \boldsymbol{e}_{m,i}= \frac{1}{|\mathcal{N}_i|} \sum_{j \in \mathcal{N}_i} \boldsymbol{e}_{j,i}.
\end{equation}
$|\mathcal{N}_i|$ is the number of observable neighboring robots of $i$. The attention score $\alpha_{j,i}$ is calculated by feeding the query and the key to a one-layer fully-connected attention network $f_a$ followed by a softmax function,
\begin{equation}
    \hat{\alpha}_{j,i}= f_a(\boldsymbol{o}_{loc,i},\boldsymbol{e}_{m,i},\boldsymbol{k}_{j,i}),
\end{equation}
\begin{equation}
    \alpha_{j,i}= \frac{e^{\hat{\alpha}_{j,i}}}{\sum_{j \in \mathcal{N}_i} e^{\hat{\alpha}_{j,i}}}.
\end{equation}
Finally, the information of all observable neighboring robots is summarized by taking the weighted mean as follows. 
\begin{equation} \label{eq:integrated_embedding}
    \boldsymbol{e}_i= \sum_{j \in \mathcal{N}_i} \alpha_{j,i}\boldsymbol{e}_{j,i}.
\end{equation}
After concatenated with $\boldsymbol{o}_{loc,i}$, $\boldsymbol{e}_i$ is taken as input by a multi-layer perception (MLP) for further inference, as shown in Fig. \ref{fig:DACOOP_A}. All the aforementioned networks $f_e$, $f_k$, $f_a$ are trained with the MLP to maximize the accumulated rewards.

\subsection{Artificial Potential Field with Attention}
Adjustments to the inter-robot distance characterize most cooperative behaviors. For example, encirclement requires pursuers to keep away from neighboring robots \cite{fang2020cooperative}. Therefore, DACOOP learns the optimal inter-robot forces $\boldsymbol{F}_{in,i}$ to prompt the emergence of cooperation \cite{dacoop}. However, it is inappropriate to average the influence of all observed robots in the computation of inter-robot forces (see (\ref{eq:individual})) as the cooperation potential of neighboring robots varies. Therefore, the proposed APF-A weights the influence of neighboring robots according to the learned attention scores when computing inter-robot forces, 
\begin{equation}
    \label{eq:ABAPF}
    \boldsymbol{F}_{in,i}=\sum_{\substack{j\neq i, j\in \mathcal{N}_i}}\alpha_{j,i}(0.5-\frac{\lambda}{d_{j,i}})\frac{\boldsymbol{p}_j-\boldsymbol{p}_i}{d_{j,i}}.
\end{equation}
The intuition behind (\ref{eq:ABAPF}) is that the neighboring robots with large attention scores usually have great potential to cooperate. Learning to adjust the distance from them is more promising for the emergence of cooperation.

\subsection{KL Divergence Regularization}
Since the behavior rules of APF-A, \emph{i.e.} (\ref{eq:ABAPF}), depend on attention scores that are updated online, the state transition probability $p(s'|s,\lambda,\eta)$ is not stationary, presenting learning stability challenges especially when replay buffer techniques are employed. Therefore, it is significant to prevent overlarge update steps of attention scores.

Denote the attention score vector of pursuer $i$ as $\boldsymbol{\alpha}_i=[\alpha_{1,i},\alpha_{2,i},\cdots,\alpha_{|\mathcal{N}_i|,i}]$. It can be treated as a categorical distribution that indicates the probability of which neighboring robot is significant. Therefore, it is possible to regulate the update of APF-A's behavior rules via minimizing the KL divergence of $\boldsymbol{\alpha}_i$ at adjacent training steps. Since calculating such KL divergence requires additional memory to store {Q networks} at the previous training step, the target network, which is introduced by DQN \cite{mnih2015human}, is employed to provide the reference distribution instead. Specifically, the KL divergence regularization is defined as 
\begin{equation}
    D_{KL}(\boldsymbol{\alpha}^-_i||\boldsymbol{\alpha}_i)=\sum_{j=1}^{|\mathcal{N}_i|} \alpha^-_{j,i} \log \frac{\alpha^-_{j,i}}{\alpha_{j,i}},
\end{equation}
where $\alpha^-_{j,i}$ is the attention score calculated by the target network. The loss function of DACOOP-A is, therefore,
\begin{equation}
    L=L_{D3QN}+c_{KL}D_{KL}(\boldsymbol{\alpha}^-_i||\boldsymbol{\alpha}_i),
\end{equation}
where $c_{KL}$ is a hyperparameter and $L_{D3QN}=[Q(s,a)-(r+\gamma \max_a Q(s',a))]^2$.

\section{Results} \label{sec:results}
In this section, the improvement of data efficiency and generalization of DACOOP-A is demonstrated via numerical simulations\footnote{Please refer to https://github.com/Zero8319/DACOOP-A for attached codes and videos.}. The feasibility of learned policies is then evaluated in physical multi-robot systems. Besides, three research questions are investigated in subsection \ref{sec:attention_effects}, \ref{sec:ABAPF_effects}, and \ref{sec:KL_effects}, respectively. First, who is attended to in the pursuit process? Second, does APF-A provide better behavior rules? Third, does KL divergence regularization stabilize the learning?

\subsection{Training settings} \label{sec:training_settings}
\begin{figure}
    \centering
    \subfloat[Training]{
    \includegraphics[width=0.15\textwidth]{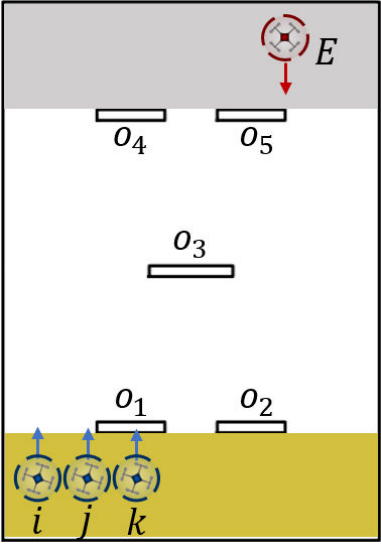}}
    \subfloat[Boundary-only]{
    \includegraphics[width=0.15\textwidth]{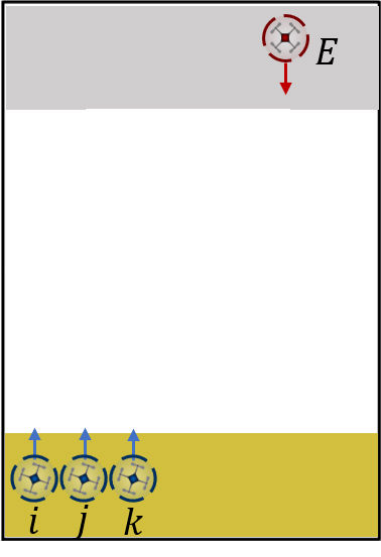}}
    \subfloat[Circle]{
    \includegraphics[width=0.15\textwidth]{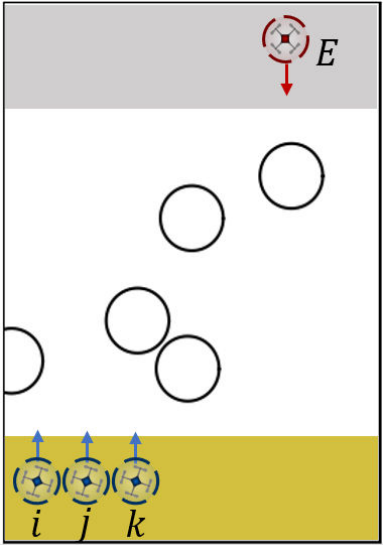}}
    \caption{The pursuit arenas. The evader $E$ is initialized randomly in the grey region while pursuers $i,j,k$ in the yellow region. (a) The training arena where $o_1,o_2,o_3,o_4,o_5$ are obstacles. (b) The validation arena where obstacles are removed beside the boundary. (c) The validation arena where five circular obstacles are generated with random positions.}
    \label{fig:arena}
\end{figure}

The pursuit arena is shown in Fig. \ref{fig:arena}(a). The problem setting parameters are listed in TABLE \ref{tab:settings}. {The escape policy of the evader is adapted from \cite{janosov2017group}. It is a force-based method. Pursuers and obstacles repulse the evader via defining multiple forces similar to (\ref{eq:attract}) and (\ref{eq:repulse}). The resultant force $\boldsymbol{F}_t$ is employed to guide the evader. The wall following rules are introduced to help the evader move along the surface of obstacles when it is in between pursuers and obstacles. As \cite{janosov2017group} does, a slip rule is employed to help the evader slip through the gap between pursuers when encircled. Totally, the escape policy is $[\boldsymbol{F}_t \vee \text{wall\_following}] \vee slip$, where $\vee$ denotes \emph{or}. For more information on the escape policy, please consult \cite{zhang2022policy} and the attached code.}

The parameter sharing techniques \cite{gupta2017cooperative} and the robust RL algorithm D3QN \cite{wang2016dueling} are employed to train the pursuit policies. The action space is 24 pairs of APF-A parameters $(\lambda,\eta)$ that are the Cartesian product of 8 $\lambda$ candidates and 3 $\eta$ candidates. Specifically, $\lambda$ candidates linearly range from 0 to $\lambda_{max}$ that is a hyperparameter, while $\eta$ candidates are chosen empirically\footnote{{We evaluate the value of $\eta_{min}$ that could turn the pursuers into the wall following mode when they are very close to obstacles. We then choose $\{\eta_{min},10\eta_{min},100\eta_{min}\}$ as $\eta$ candidates.}}. The reward function consists {of} three terms, $r=r_{main}+r_{col}+r_{app}$, where $r_{main}$ gives a reward of 20 when the evader is captured. $r_{col}$ gives a punishment of -20 if collisions occur. $r_{app}$ is a reward shaping term \cite{ng1999policy} that awards pursuers when they approach the evader. 

In addition to the proposed algorithm DACOOP-A, several benchmark algorithms are trained as follows.

\begin{itemize}
       \item {ME \cite{huttenrauch2019deep}. It combines D3QN with mean embedding. The action space is 24 discretized expected headings.  }
       \item {D3QN-att. It combines D3QN with attention. The implementation details of attention are the same as DACOOP-A. The action space is 24 discretized expected headings.}
        \item DACOOP \cite{dacoop}. The action space is 24 parameter pairs $(\lambda,\eta)$ for vanilla APF. Mean embedding is also used.
        \item {MAAC \cite{iqbal2019actor}. MAPPO is selected as the backbone \cite{yu2022surprising}. The attention mechanisms are employed to synthesize the observations of pursuers in the centralized critic.}
        \item {No-RL \cite{janosov2017group}. It is a non-learning method whose hyperparameters are tuned via the evolutionary algorithm. Since it is difficult for this non-learning method to accomplish the original task, the pursuit problem is simplified so that episodes end when \emph{any} pursuer captures the evader.}
        \item No-KL. It is an ablation study by removing KL divergence regularization from DACOOP-A. 
        \item {DACOOP-att}. It is an ablation study by removing both KL divergence regularization and APF-A from DACOOP-A. It implies {DACOOP-att} is the pure combination of DACOOP and the attention mechanisms.
\end{itemize} 
{All algorithms are performed with five random seeds.} RL hyperparameters and their values are listed in TABLE \ref{tab:settings}. Note that the basic hyperparameters, \emph{e.g.}, learning rate, are tuned for ME and then employed by D3QN-att, DACOOP, and DACOOP-A without tuning for a fair comparison. Similarly, we tune $\lambda_{max}$ for DACOOP and then directly adopt the results in DACOOP-A. Only $c_{KL}$ is tuned for DACOOP-A. {The hyperparameters of MAAC are tuned independently.}

\begin{table}
    \centering
    \caption{Problem settings and RL hyperparameters}
    \begin{tabular}{cc|cc}
        \toprule
        \label{tab:settings}
        Parameters & Values & Parameters & Values\\
        \midrule
        $N$ & 3 & Arena size & $3.6 \; m \times 5 \; m$ \\
        $d_c$ & $0.2 \; m$ & $d_s$ & $0.1 \; m$ \\
        $d_p$ & $2 \; m$ & Step size  & $0.1 \; s$ \\
        $v_P$ & $0.3 \; m/s$ & $v_E$ & $0.6 \; m/s$\\
        $t_{max}$ & $100 \; s$ & $\rho$ & $0.8 \; m$ \\
        Learning rate & 3e-5 & Memory size & 1e6\\
         Discount factor & 0.99 & $\lambda_{max}$ & 4000\\
         Maximal episode & 4000 & final exploration episode & 2000\\
         Minibatch size & 128 & $c_{KL}$ & 0.05\\
        \bottomrule
    \end{tabular}
\end{table}

\subsection{Data Efficiency and Generalization Ability} \label{sec:data_efficiency}
\begin{figure}
    \centering
    \includegraphics[width=0.48\textwidth]{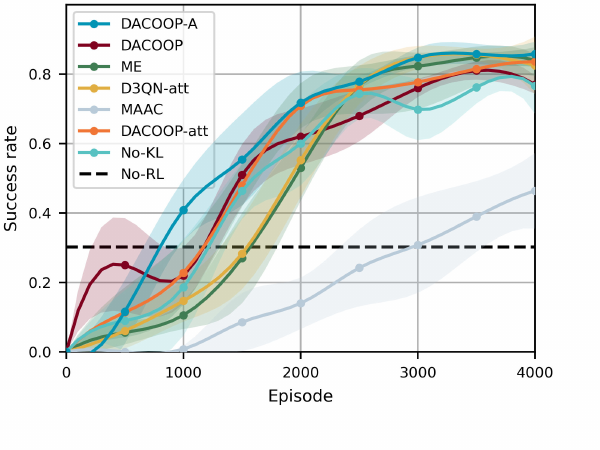}
    \caption{{The learning curves of different algorithms. All results are averaged over 1000 validation episodes and five random seeds. The shaded areas indicate the 95$\%$ confidence interval.}}
    \label{fig:learning_curves}
\end{figure}

The learning curves are demonstrated in Fig. \ref{fig:learning_curves}, and the relative statistical results are listed in TABLE \ref{tab:AUC}. The area under the learning curve (AUC), \emph{i.e.} the mean success rate achieved at nine checkpoints in Fig. \ref{fig:learning_curves}, is used as the metric of data efficiency as \cite{ostrovski2017count}. From TABLE \ref{tab:AUC}, it is observed that DACOOP-based algorithms outperform all baselines in terms of data efficiency due to the introduction of APF. {DACOOP-A achieves the best AUC of 0.57 as it employs attention mechanisms to prompt information processing and refine the behavior rules. The poor performance of No-KL indicates the importance of KL regularization in alleviating the non-stationarity issue. MAAC does not obtain a satisfactory AUC because the on-policy RL methods are more data inefficient than off-policy ones in most scenarios. The learned policies' maximal success rate (MSR) is also listed in TABLE \ref{tab:AUC}.} Since the action space of DACOOP is not complete, \emph{i.e.}, there may be some expected headings unavailable no matter what value $(\lambda,\eta)$ takes, the MSR of DACOOP is inferior to that of ME. However, DACOOP-A alleviates such issues by refining vanilla APF with attention scores, resulting in competitive asymptotic performance while maintaining better AUC. {In addition, the collision rate in the training process (CRTP) of DACOOP-based algorithms is much lower than baselines as APF(-A) provides more knowledge in collision avoidance, which is significant for the safety issue in the physical systems training\cite{dalal2018safe}.}

\begin{table}
    \centering
    \caption{{The mean and standard deviation of the learning results across different random seeds}}
    \begin{tabular}{c|ccc}
        \toprule
        \label{tab:AUC}
        & {AUC}  &{MSR} &{CRTP}\\
        \midrule
        {DACOOP-A}   & {$\boldsymbol{0.57\pm0.03}$}    &{$\boldsymbol{0.87\pm0.03}$} &{$\boldsymbol{0.32\pm0.01}$}\\
        {DACOOP}     & {$0.51\pm0.05$}    &{$0.81\pm0.02$} &{$0.35\pm0.02$}\\
        {ME}         & {$0.47\pm0.02$}    &{$0.86\pm0.02$} &{$0.48\pm0.01$}\\
        {D3QN-att}   & {$0.48\pm0.03$}    &{$\boldsymbol{0.87\pm0.03}$} & {$0.47\pm0.01$}\\
        {MAAC}       & {$0.18\pm0.06$}    &{$0.52\pm0.13$} &{$0.73\pm0.04$}\\
        {DACOOP-att} & {$0.54\pm0.05$}    &{$0.85\pm0.04$} &{$0.33\pm0.03$}\\
        {No-KL}      & {$0.48\pm0.04$}    &{$0.80\pm0.06$} &{$0.35\pm0.01$} \\
        \bottomrule
    \end{tabular}
\end{table}

\begin{figure}
    \centering
    \subfloat[Number]{
    \includegraphics[width=0.48\textwidth]{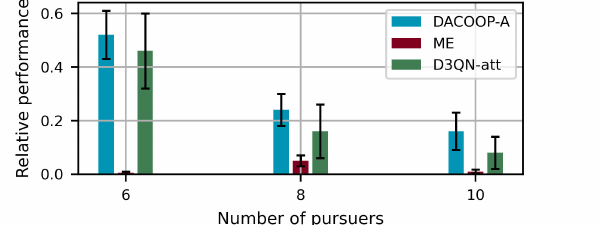}} \\
    \subfloat[Arena]{
    \includegraphics[width=0.48\textwidth]{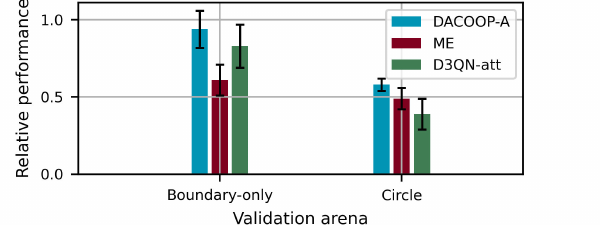}}
    \caption{The generalization performance of different algorithms. The relative performance is defined as the ratio of the success rate in the validation scenarios to that in the training scenarios. {The error bars indicate the standard deviation across different random seeds.}}
    \label{fig:generalization}
\end{figure}

The generalization performance of different algorithms is shown in Fig. \ref{fig:generalization}. It is observed from \ref{fig:generalization}(a) that DACOOP-A is more robust than {ME and D3QN-att} when the system size changes. To investigate the underlying reasons, we attempt to measure the variation in observation embedding results when algorithms are deployed in validation scenarios. Firstly, we collect $2 \times 10^6$ environment states for each algorithm via uniformly sampling robots' positions and headings. Thereinto, $10^6$ states are 3-pursuer scenarios while the other $10^6$ states are {10-pursuer scenarios}. The integrated embedding $\boldsymbol{e}_i$ is calculated for all pursuers at all states. Note that $\boldsymbol{e}_i$ is the mean embeddings for ME, while is calculated according to (\ref{eq:integrated_embedding}) for DACOOP-A and {D3QN-att}. Two matrices $A\in \mathcal{R}^{3e6 \times 128}$ and $B \in \mathcal{R}^{6e6 \times 128}$ are acquired for each algorithm, where $A$ {consists} of all integrated embedding results $\boldsymbol{e}_i$ in 3-pursuer scenarios while $B$ consists of that in {10-pursuer scenarios}. 128 is the length of embedding vectors. Considering each matrix as a point set in a 128-dimensional space, we use the Hausdorff distance to measure {the difference between $A$ and each point in $B$. The results are averaged over all points in $B$ and denoted as AHD (averaged Hausdorff distance) in TABLE \ref{tab:variation}.} It could be observed that the integrated embeddings $\boldsymbol{e}_i$ of ME change dramatically when they are deployed in systems with different sizes. So, strange state representations are the major impediment to generalization. In comparison, the attention module of DACOOP-A preserves significant information while suppressing redundant information, making $\boldsymbol{e}_i$ more invariant. {Although the AHD of D3QN-att is similar to that of DACOOP-A, its success rate is still much lower. The reason is that the optimal policy changes when the system size differs. However, similar observation embeddings result in similar actions in D3QN-att. In comparison, the behavior rules of APF-A depend on the system size, which is promising to provide the desirable adaption based on invariant observation embeddings.\footnote{{Note that it should not be expected that agents perform well with unseen observations. The generalization ability of DACOOP-A derives from the adaptation of the policy (Q network + APF-A) instead of the variance of the observation embeddings.}}} Fig. \ref{fig:generalization}(b) shows the generalization ability of each algorithm in validation arenas. The Boundary-only arena provides more free space for both pursuers and the evader, making cooperation more significant than obstacle avoidance for pursuers (see Fig. \ref{fig:arena}(b)). Therefore, the better generalization of DACOOP-A in this scenario demonstrates that more sophisticated and intelligent cooperative behaviors are learned due to the direct regulation of the distance from significant neighboring robots. The Circle arena employs different obstacles as shown in Fig. \ref{fig:arena}(c). The better performance of DACOOP-A in this arena verifies the contributions of wall following rules to obstacle avoidance (see Fig. \ref{fig:generalization}(b)).

\begin{table}
    \centering
    \caption{{The mean and standard deviation of AHD across different random seeds}}
    \begin{tabular}{c|ccc}
        \toprule
        \label{tab:variation}
        & {DACOOP-A} & {ME} & {D3QN-att}\\
        \midrule
        {AHD}  & {$\boldsymbol{0.235\pm0.018}$} & {$0.587\pm0.031$} &{$0.243\pm0.046$}\\
        \bottomrule
    \end{tabular}
\end{table}

\subsection{Effects of Attention} \label{sec:attention_effects}
To investigate who is attended to in the pursuit process, 3000 episodes are rolled out with policies learned by DACOOP-A. For each pursuer $i$, we measure the influence of neighboring robots on $i$'s state value by $|V(s_{i,-j})-V(s_i)|$ and $|V(s_{i,-k})-V(s_i)|$, where $V(\cdot)$ is the state value function. $s_i$ is the local observations of $i$. $s_{i,-j}$ denotes removing $j$ from $i$' observations, while $s_{i,-k}$ denotes removing $k$. Let $E_1$ denote the event satisfying $$\big(|V(s_{i,-j})-V(s_i)|-|V(s_{i,-k})-V(s_i)|\big)\big(\alpha_{j,i}-\alpha_{k,i}\big)>0$$ while $E_2$ denotes events where $$\big(|V(s_{i,-j})-V(s_i)|-|V(s_{i,-k})-V(s_i)|\big)\big(\alpha_{j,i}-\alpha_{k,i}\big) \leq 0.$$ Their frequencies are calculated over all trajectories collected in the aforementioned 3000 episodes. {Since state values depend on the policy, five critics trained by ME with different random seeds are employed to evaluate $V(\cdot)$ for justice}\footnote{{The dueling network in D3QN has two streams to separately estimate the scalar state value and the advantages for each action \cite{wang2016dueling}. Therefore, we take the output of the first stream as the state value in this work.}}. The results shown in TABLE \ref{tab:event} demonstrate that all critics consistently think $E_1$ is more frequent than $E_2$. It implies that the neighboring robots influential on state values are more likely to be attended to. Given the reward function used in this paper, it could be concluded that pursuers attend to neighboring robots mainly for collision avoidance and cooperation.

\begin{table}
    \centering
    \caption{The Frequency of Corresponding Events}
    \begin{tabular}{c|ccccc}
        \toprule
        \label{tab:event}
        Event & critic1 & critic2 & critic3 & {critic4} & {critic5}\\
        \midrule
        $E_1$ & \textbf{75.4}\% & \textbf{73.7}\%  & \textbf{72.5}\%  & {\textbf{73.2}\%}  & {\textbf{74.5}\%} \\
        $E_2$ &  24.6\% & 26.3\%  & 27.5\%  & {26.8\%}  & {25.5\%} \\
        \bottomrule
    \end{tabular}
\end{table}

\subsection{Effects of APF-A} \label{sec:ABAPF_effects}
The magnitude of inter-robot forces increases as inter-robot distance decreases in APF. So the resultant forces $\boldsymbol{F}_i$ of APF-A are similar to that of APF if closer robots are attended to. To distinguish the effects of APF-A, the environment states satisfying the following conditions are selected from the aforementioned 3000 episodes.
\begin{itemize}
    \item {The attention scores $\alpha_{j,i}>\alpha_0$ while $\alpha_{k,i}<1-\alpha_0$, where $\alpha_0>0.5$. It implies that $i$ attends to $j$ while $k$ is neglected by $i$.}
    \item The distance between $i$ and $E$ is less than $5d_c$. It implies that $i$ is in a situation where cooperation is important.
    \item The distance between $i$ and $j$ is larger than that between $i$ and $k$, meaning the distant robot is attended to. 
\end{itemize}

\begin{figure}
    \centering
    \subfloat[]{
    \includegraphics[width=0.23\textwidth]{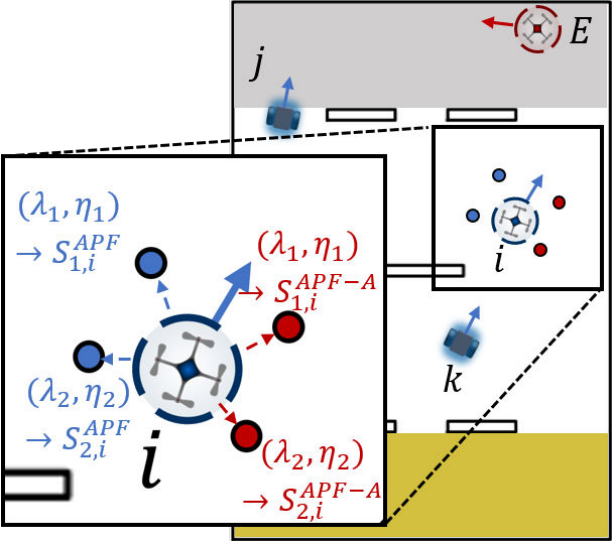}}
    \subfloat[]{
    \includegraphics[width=0.25\textwidth]{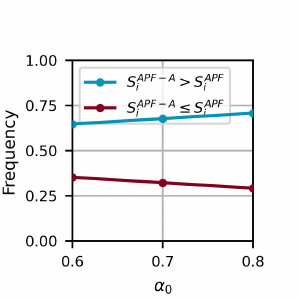}}
    \caption{(a) Demonstrations of the evaluation of $S_i^{APF}$ and $S_i^{APF\text{-}A}$. Here taking $|\mathcal{A}|=2$ for example. Blue (red) circles denote the resultant positions of pursuer $i$ if it moves according to APF (APF-A). The formation score $S_i$ is evaluated for each resultant position. $S^{APF}_i$ is the sum of formation scores evaluated at blue circles, while $S^{APF\text{-}A}_i$ is that at red circles. (b) {The frequency of the events $S^{APF\text{-}A}_i>S^{APF}_i$ and $S^{APF\text{-}A}_i \leq S^{APF}_i$ when $\alpha_0$ takes different values.}}
    \label{fig:ABAPF_effects}
\end{figure}

Similar to \cite{de2021decentralized}, the formation score is defined for pursuer $i$ to evaluate the potential for encirclement at a certain state,
\begin{equation}
    S_i=\sum_{j=1,i \neq j}^N (\boldsymbol{p}_i-\boldsymbol{p}_E)(\boldsymbol{p}_j-\boldsymbol{p}_E)^T.
\end{equation}
As shown in Fig. \ref{fig:ABAPF_effects}(a), the usefulness of behavior rules at a certain state $s$ is measured by 
\begin{equation}
    S_i^{APF}=\sum_{a=1}^{|\mathcal{A}|} S^{APF}_{a,i}, \quad
    S_i^{APF\text{-}A}=\sum_{a=1}^{|\mathcal{A}|} S^{APF\text{-}A}_{a,i},
\end{equation}
where $S^{APF}_{a,i}$ is the formation score $S_i$ of the state transited from $s$ by moving pursuer $i$ according to APF with $a$-th parameter pair. $S^{APF\text{-}A}_{a,i}$ is that by moving according to APF-A instead. Note that although the 24 parameter pairs $(\lambda,\eta)$ are the same for APF-A and APF in this paper, the expected headings differ due to different behavior rules. Both $S_i^{APF}$ and $S_i^{APF\text{-}A}$ are evaluated for all selected states. {The proportion of states with $S_i^{APF\text{-}A}>S_i^{APF}$ is much more than that with $S_i^{APF\text{-}A} \leq S_i^{APF}$ no matter what value $\alpha_0$ takes as shown in Fig. \ref{fig:ABAPF_effects}(b).} It suggests that APF-A provides candidate headings with better quality due to the direct regulation of the distance from significant neighbors.

 \begin{figure}
    \centering
    \includegraphics[width=0.48\textwidth]{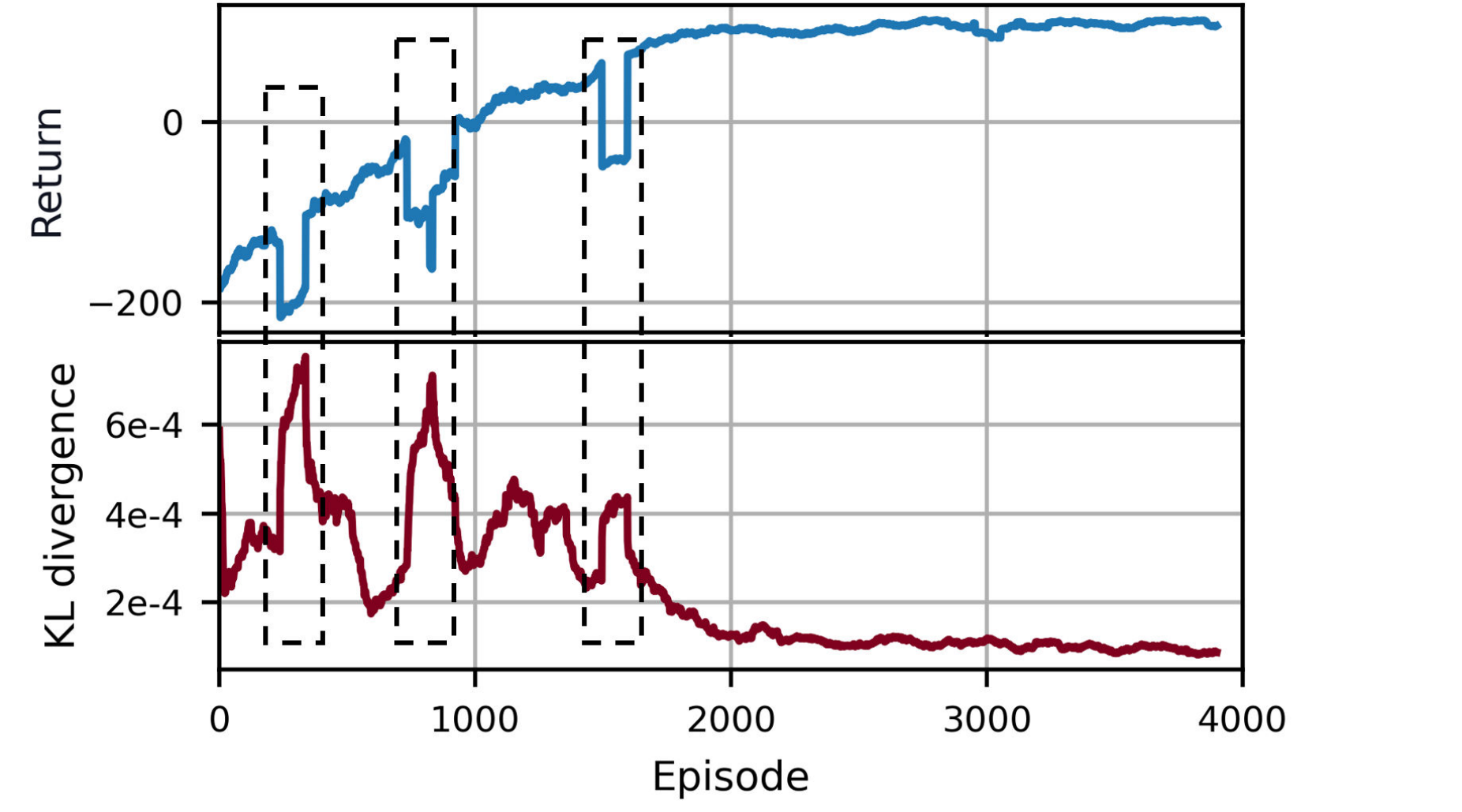}
    \caption{An example of the learning curves of DACOOP-A. The rise of KL divergence of attention scores always induces drops in accumulated rewards.}
    \label{fig:demonstrations}
\end{figure}

\subsection{Effects of KL Divergence Regularization} \label{sec:KL_effects}
The non-stationarity issue derived from evolutionary behavior rules usually induces unstable learning characterized by sudden drops of accumulated rewards, as shown in Fig. \ref{fig:demonstrations}. To quantify the underlying relations, we record the KL divergence of attention scores $D_j$ and accumulated rewards $R_j$ at each training step $j$. The result is a list, $\{(D_j,R_j)\}_{j=1}^{T_{max}}$, where $T_{max}$ is the maximum training step. Then, their gradients $\{(\nabla D_j,\nabla R_j)\}_{j=1}^{T_{max}}$ are calculated via the first-order forward difference. To make the result significant, we evaluate the Pearson correlation coefficient between $\nabla D$ and $\nabla R$ over data points $\{(\nabla D_j,\nabla R_j)\}_j$ that meet $\nabla D_j>c_{grad}$, where $c_{grad}$ is a threshold. The results shown in TABLE \ref{tab:kl} show that the Pearson correlation coefficient is less than -0.4 if $c_{grad}$ is greater than 1e-5. The existence of moderate negative correlations suggests that an increase in KL divergence usually induces a decrease in the accumulated rewards. Therefore, avoiding an over-large divergence of KL attention scores in the learning process is significant. To investigate whether KL divergence regularization stabilizes the learning process, the overlarge $\nabla D_j$ amount is evaluated for DACOOP-A and No-KL, respectively. It can be observed from TABLE \ref{tab:kl} that DACOOP-A always obtains fewer data points satisfying $\nabla D_j>c_{grad}$, which explains why KL divergence regularization enables greater data efficiency.

\begin{table}
    \centering
    \caption{Statistic Results in Selected Data Points}
    \begin{tabular}{c|cccc}
        \toprule
        \label{tab:kl}
        $c_{grad}$ & 1e-5 & 3e-5 & 5e-5 & 7e-5\\
        \midrule
        $Pearson(\nabla D,\nabla R)$ & \textbf{-0.33} & \textbf{-0.43} & \textbf{-0.46} & \textbf{-0.50}\\
        P-value  & 1e-19 & 5e-6 & 1e-3 & 0.01\\
        Amount of data points (No-KL)  & 195 & 22 & 14 & 9\\
        Amount of data points (DACOOP-A)  & \textbf{149} & \textbf{20} & \textbf{6} & \textbf{3}\\
        \bottomrule
    \end{tabular}
\end{table}

 \begin{figure}
    \centering
    \subfloat[]{
    \includegraphics[width=0.15\textwidth,height=0.224\textwidth]{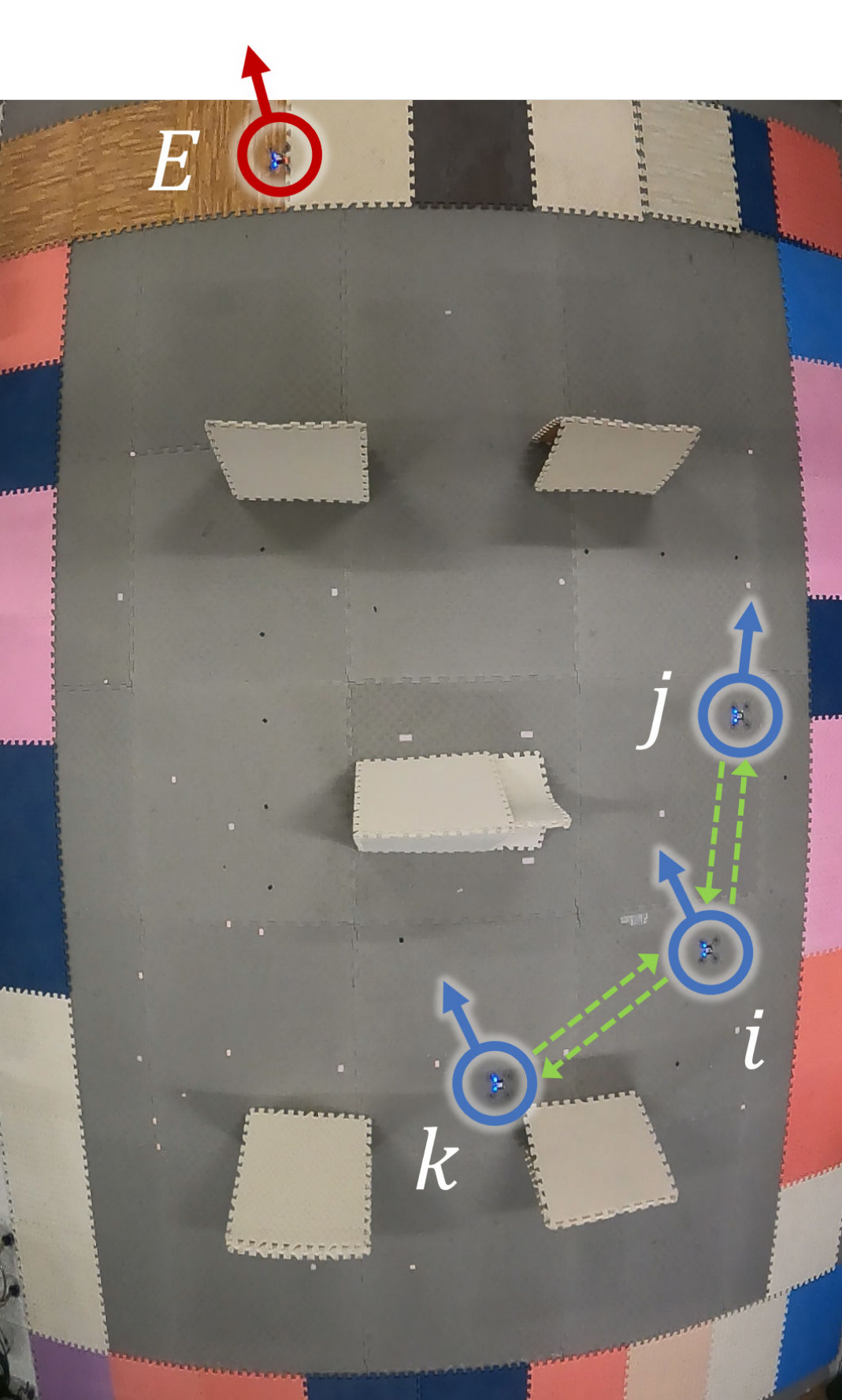}}
    \subfloat[]{
    \includegraphics[width=0.15\textwidth,height=0.2095\textwidth]{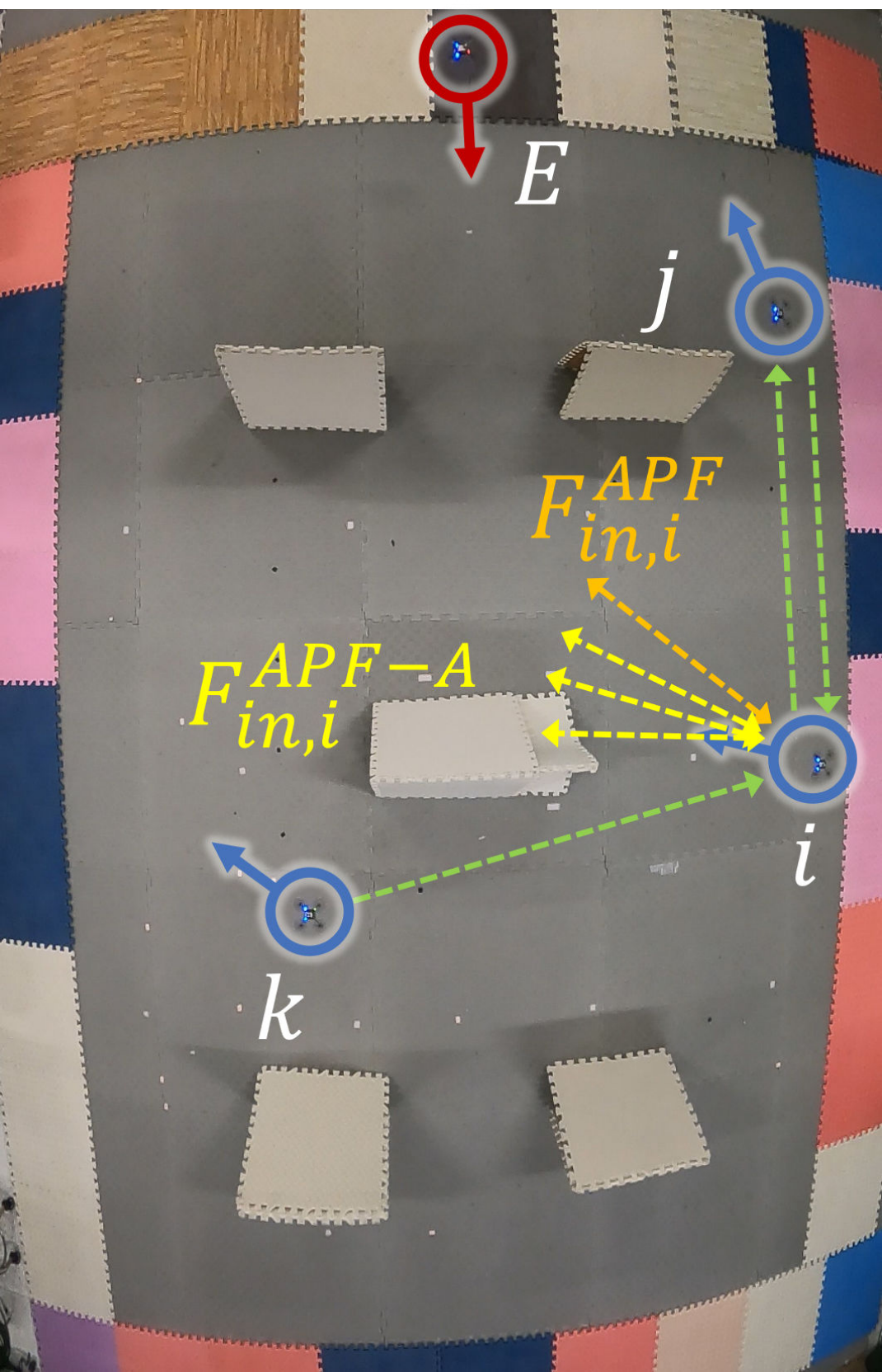}}
    \subfloat[]{
    \includegraphics[width=0.15\textwidth,height=0.2095\textwidth]{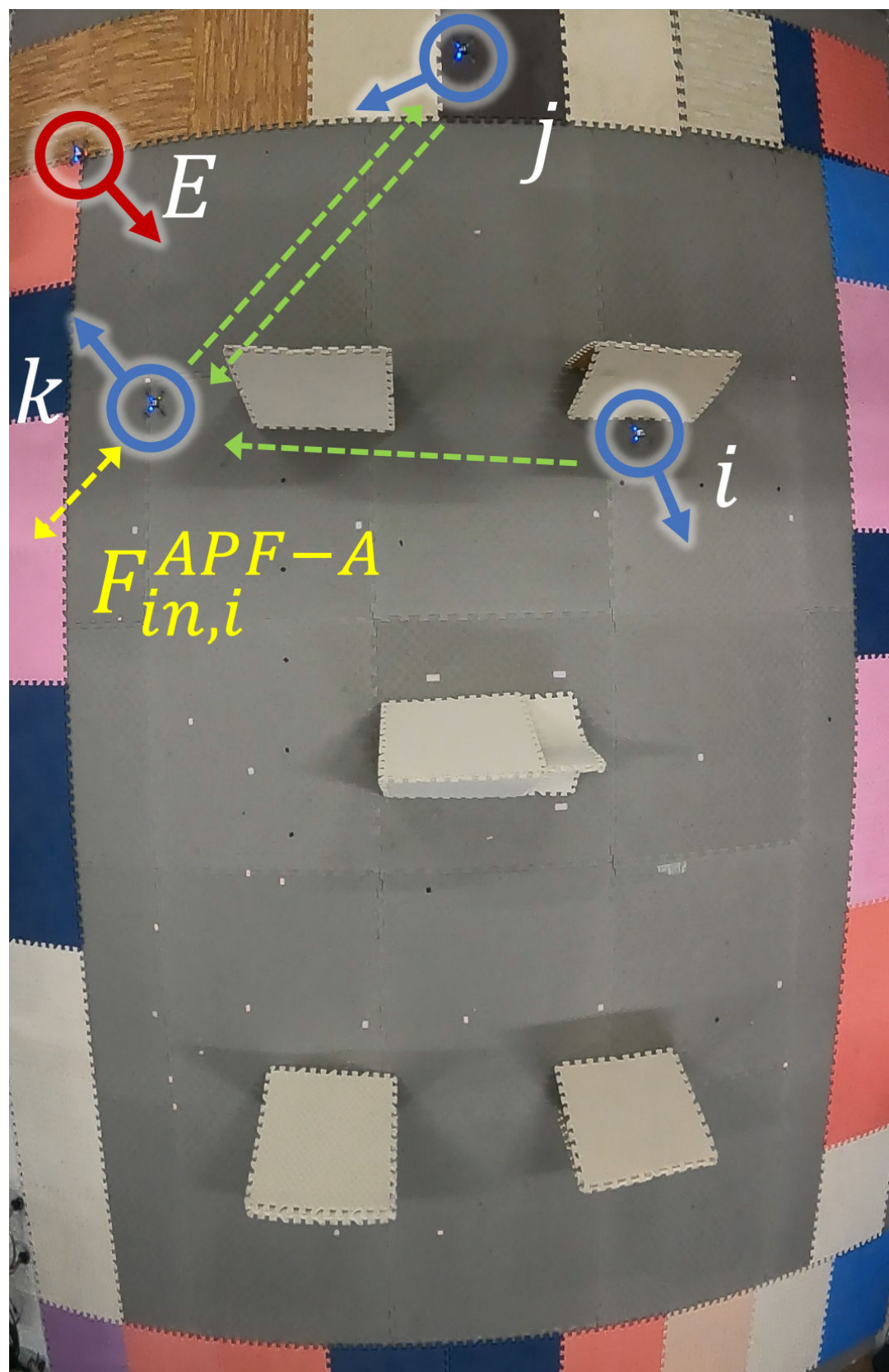}}
    \caption{The snapshots of real-world experiments. The green arrows point from ego-quadrotors to neighbors who are attended to. The yellow and orange arrows denote $F_{in,i}$ evaluated by APF-A and APF, respectively. }
    \label{fig:experiment}
\end{figure}

\subsection{Physical Experiments} \label{sec:physical_experiments}

The learned policies of DACOOP-A are deployed in multi-quadrotor systems {Crazyflie}\footnote{{https://www.bitcraze.io/products/crazyflie-2-1/}} directly. The positions and orientations of quadrotors are measured by the motion capture system OptiTrack. The key snapshots are shown in Fig. \ref{fig:experiment}. In Fig. \ref{fig:experiment}(a), pursuers are distant from the evader. Since their main concern in this state is approaching the evader safely, all pursuers attend to the nearest neighboring robots for collision avoidance. Rather than giving similar attention to $j$ and $k$ in Fig. \ref{fig:experiment}(a), pursuer $i$ switches to attend to $j$ in Fig. \ref{fig:experiment}(b) because $j$ is adjacent to the evader and has more potential to cooperate with. Note that the distance between $i$ and $j$ is similar to that between $i$ and $k$ in Fig. \ref{fig:experiment}(b). It implies that the inter-robot forces evaluated by APF always align with the angular bisector because the forces exerted by neighboring robots are symmetric (see Fig. \ref{fig:experiment}(b)). In comparison, APF-A weights the influence of neighboring robots according to attention scores in evaluating inter-robot forces, providing diverse candidate expected headings even if $i$ is just between two neighboring robots. In Fig. \ref{fig:experiment}(c), pursuer $k$ attends to $j$ since the encirclement they form is the necessity of successful capture. To verify the significance of directly regulating the distance between $k$ and $j$, the formation scores $S^{APF}_{a,k}$ and $S^{APF\text{-}A}_{a,k}$ are evaluated for 24 parameters pairs ($\lambda_a,\eta_a$) in Fig. \ref{fig:experiment}(c). The results show that only 1 out of 24 parameter pairs satisfies $S^{APF}_{a,k}>S^{APF\text{-}A}_{a,k}$, proving that taking the insignificant neighboring robots $i$ into consider hinders the formation of encirclement. By APF-A, pursuer $k$ cuts off all possible escape routes of the evader in Fig. \ref{fig:experiment}(c) and then successfully captures it.

\section{Conclusions} \label{sec:conclusions}
This paper proposes a multi-robot pursuit algorithm named DACOOP-A by empowering vanilla RL with APF and attention mechanisms. Simulation results demonstrate better data efficiency, competitive asymptotic performance, and lower collision rate in the training process of DACOOP-A. It is also verified that DACOOP-A has greater generalization ability regarding system size and pursuit arena. Further analysis demonstrates that neighboring robots who influence state values are likely to be attended to. In addition, APF-A is proven to provide evolutionary behavior rules that are more promising for encirclement. Simulation results also show that a regularization could alleviate the non-stationarity issue by avoiding overlarge gradients of KL divergence of attention scores in the learning process. Physical experiments verify the feasibility of directly deploying the learned policies in real-world multi-robot systems. However, the action space is not complete in DACOOP-A. Enabling robots to select all possible headings will be considered in future works. {Besides, extending the attention mechanisms to multiple obstacles or evaders is also an interesting direction.}


\bibliography{references}
\bibliographystyle{IEEEtran}
\end{document}